\documentclass{article}

\usepackage{PRIMEarxiv}

\usepackage[utf8]{inputenc} % allow utf-8 input
\usepackage[T1]{fontenc}    % use 8-bit T1 fonts
\usepackage{hyperref}       % hyperlinks
\usepackage{url}            % simple URL typesetting
\usepackage{booktabs}       % professional-quality tables
\usepackage{amsfonts}       % blackboard math symbols
\usepackage{nicefrac}       % compact symbols for 1/2, etc.
\usepackage{microtype}      % microtypography
\usepackage{amsmath}        % for \text and math environments
\usepackage{lipsum}
\usepackage{fancyhdr}       % header
\usepackage{graphicx}       % graphics
\graphicspath{{media/}}     % organize your images and other figures under media/ folder

\title{Multimodal Alignment with Cross-Attentive GRUs for Fine-Grained Video Understanding

}

\author{
  NAMHO KIM \\
  Korean Broadcasting System(KBS) \\
  Seoul, Korea\\
  \texttt{namho96@kbs.co.kr} \\
      \And
  JUN-HWA KIM \\
  Department of Artificial Intelligence \\
  Konyang University \\
  Daejeon, Korea\\
  \texttt{junhwakim@konyang.ac.kr} \\
}

\begin{document}
\maketitle

\begin{abstract}
Fine-grained video classification requires understanding complex spatio-temporal and semantic cues that often exceed the capacity of a single modality. In this paper, we propose a multimodal framework that fuses video, image, and text representations using GRU-based sequence encoders and cross-modal attention mechanisms. The model is trained using a combination of classification or regression loss, depending on the task, and is further regularized through feature-level augmentation and autoencoding techniques. To evaluate the generality of our framework, we conduct experiments on two challenging benchmarks: the DVD dataset for real-world violence detection and the Aff-Wild2 dataset for valence-arousal estimation. Our results demonstrate that the proposed fusion strategy significantly outperforms unimodal baselines, with cross-attention and feature augmentation contributing notably to robustness and performance.
\end{abstract}

% keywords can be removed

\section{Introduction}
This study aims to enhance the intricate scene analysis capabilities required for demanding video understanding tasks such as violence detection and affective state estimation. While vision-based architectures, including convolutional neural networks (CNNs) and transformer models, have achieved significant progress in visual recognition, they often fall short in capturing subtle semantic distinctions, particularly in fine-grained or emotionally rich scenarios. These limitations are further exacerbated in video settings, where spatio-temporal dependencies must be jointly modeled across multiple frames.

To mitigate such limitations, prior research has explored the integration of multimodal information—including vision, text, and audio—to provide complementary cues for comprehensive scene understanding. For instance, approaches like CLIP align visual and textual embeddings to boost classification or retrieval accuracy in joint vision-language tasks \cite{radford2021clip}. Similarly, challenges in affective behavior analysis, such as ABAW \cite{Kollias2025, kolliasadvancements,kollias20247th, kollias20246th, kollias2024distribution, kollias2023abaw2, kollias2023multi, kollias2023abaw, kollias2022abaw, kollias2021analysing, kollias2020analysing, kollias2021distribution, kollias2021affect, kollias2019expression, kollias2019face, kollias2019deep}, have demonstrated the utility of combining facial expressions, audio, and contextual information for robust emotion recognition \cite{kollias2023abaw, Kollias2025}. Despite these advancements, many existing models frequently rely on complex transformer-based fusion mechanisms, which, while effective, often incur high computational costs and complexity during both training and inference.

In this paper, we propose a robust and efficient multimodal fusion framework that incorporates three parallel modalities: video segments capturing motion dynamics, sampled image frames offering rich spatial detail, and textual captions extracted from keyframes that convey explicit semantic information. Each modality is processed through a frozen pre-trained encoder—specifically, a 3D CNN for video, a vision transformer for images, and a language encoder for text—followed by GRU-based sequence encoders to model temporal features within each respective stream.

We employ a bidirectional cross-attention mechanism to intricately integrate the modality-specific features, facilitating dynamic and context-aware interaction among video, image, and text representations. The resulting fused embedding is subsequently passed through a shallow multilayer perceptron for final classification or regression. Model training is guided by a composite loss function: cross-entropy is utilized for classification tasks (e.g., violence detection on the DVD dataset), and mean squared error is applied for regression tasks (e.g., valence-arousal estimation on the Aff-Wild2 dataset).

To validate our approach, we conduct comprehensive experiments on two challenging benchmark datasets: the DVD dataset for real-world violence detection \cite{kollias2025dvd}, and the Aff-Wild2 dataset for continuous valence-arousal prediction \cite{kollias2019expression}. Both datasets present demanding scenarios that explicitly require robust multimodal reasoning. Our experimental results consistently demonstrate that the proposed framework significantly outperforms unimodal and naive fusion baselines. Furthermore, thorough ablation studies confirm that both the cross-attention mechanism and feature-level augmentation are critical components, significantly enhancing the model's robustness and performance.

This research contributes a practical and highly effective architecture for fine-grained video understanding and paves the way for broader applications in critical domains such as public safety, healthcare, and advanced human-computer interaction.

\section{Related Work}

\subsection{Multimodal Feature Representation}

Recent advancements in affective behavior analysis and scene understanding have emphasized the importance of multimodal feature extraction from video data. In large-scale benchmarks such as the Aff-Wild2 dataset, visual, audio, and textual information are often jointly exploited to improve recognition accuracy across diverse affective tasks \cite{kollias2023abaw}. Rather than relying solely on raw video input, researchers have increasingly adopted pretrained encoders to obtain more robust and semantically meaningful features from each modality.

For visual representation, deep convolutional and transformer-based models such as ResNet \cite{he2016deep}, EfficientNet \cite{tan2021efficientnetv2}, and masked autoencoders (MAE) \cite{he2022masked} have been widely used, often trained on domain-specific datasets like facial expression corpora to improve generalization. In parallel, audio features are commonly extracted using pretrained models such as VGGish \cite{hershey2017cnn} or Wav2Vec2 \cite{baevski2020wav2vec}, which capture both low-level acoustic properties and high-level emotional tones. Textual features, derived from transcriptions or generated captions, are typically embedded using large language models such as BERT \cite{devlin2019bert}.

Multimodal setups have also expanded to incorporate three or more streams. Unlike earlier approaches that focused solely on vision or audio, recent research increasingly combines all three—visual, audio, and language—to capture the richness of human behavior and scene semantics. These multimodal embeddings form the foundation for downstream tasks including emotion regression and violence detection.

In our work, we adopt a similar philosophy by encoding video segments, image frames, and generated captions using frozen pretrained models. Unlike many prior approaches that focus on facial regions or speech alone, we combine scene-level visual features with semantic textual information to enrich the understanding of complex situations such as aggression or emotional dynamics.

\subsection{Multimodal Fusion Strategies}

Beyond feature extraction, how these multimodal representations are fused plays a crucial role in determining system performance. Early works commonly employed naive fusion strategies, such as direct concatenation followed by fully connected layers. However, such methods often struggle to model inter-modal relationships effectively.

Recent approaches increasingly adopt attention-based fusion mechanisms. Transformer-based models \cite{vaswani2017attention} use self-attention to jointly process features from multiple modalities, learning complex dependencies. While powerful, these models are often computationally intensive and require large datasets to train effectively. For temporal modeling, GRUs \cite{cho2014learning} and Temporal Convolutional Networks (TCNs) \cite{bai2018empirical} offer efficient alternatives, particularly when working with smaller datasets or deploying on resource-constrained systems.

Our approach builds on these trends by employing GRU-based sequence encoders for each modality and integrating them via bidirectional cross-attention. This design enables each modality to inform and enhance the others while maintaining modularity and computational efficiency. By doing so, we bridge the semantic and temporal gaps between modalities without the overhead of full transformer-based fusion.

\section{Methodology}

\begin{figure*}[htbp]
\centering
\includegraphics[width=\textwidth]{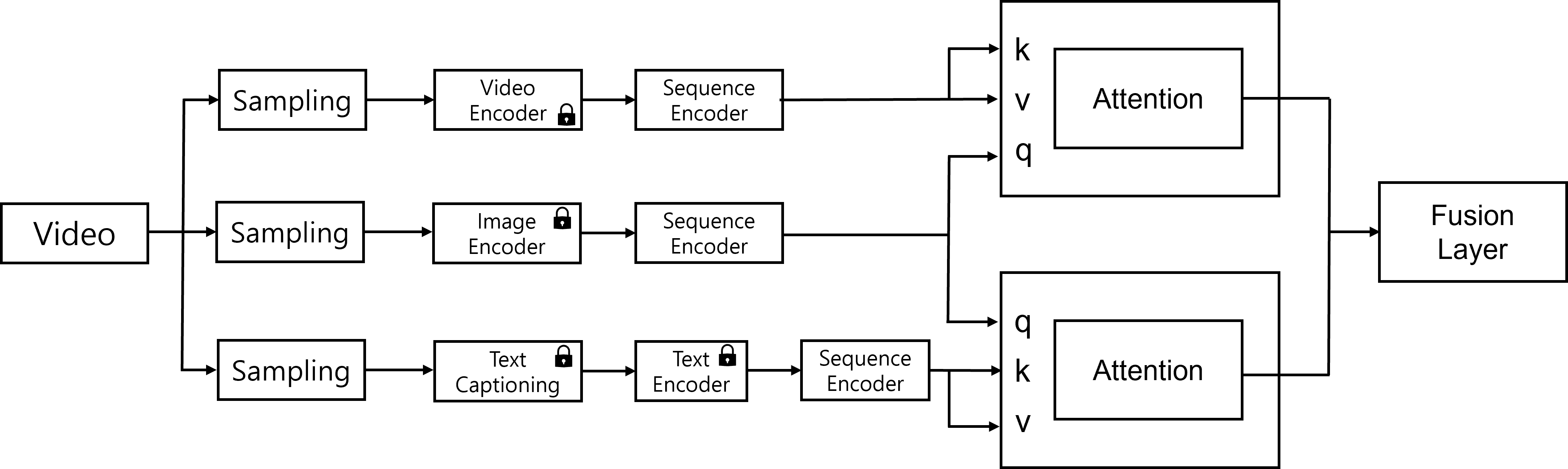}
\caption{Overall structure of the proposed network.}
\label{fig:structure}
\end{figure*}

Our proposed multimodal framework is designed to integrate video, image, and text modalities derived from a single video source. As shown in Figure \ref{fig:structure}, three modality-specific streams are sampled and processed in parallel. Each stream is passed through a pretrained encoder followed by a GRU-based sequence encoder, and the resulting embeddings are fused using cross-attention mechanisms before being passed to the final prediction head.

To ensure temporal coverage and semantic diversity, we uniformly sample 64 segments from each input video. From these, a subset of 16 or 4 samples is chosen per modality stream depending on the encoder type and data type, as described below.

\subsection{Modality-Specific Feature Extraction}

\subsubsection{Visual Motion Representation from Sampled Frames}

We use a sliding window of size $64$ over each video. From each 64-frame window, we uniformly sample $16$ frames, denoted as $\{x_i^{(v)}\}_{i=1}^{16}$. These sampled RGB frames are stacked into a tensor $X^{(v)} \in \mathbb{R}^{16 \times 3 \times H \times W}$ and input into a frozen \textbf{VideoPrism} encoder \cite{georgescu2023video} to extract high-level spatiotemporal features:
\begin{equation}
\{f^{(v)}_i\}_{i=1}^{16} = \text{VideoPrism}(X^{(v)}), \quad f^{(v)}_i \in \mathbb{R}^{d_v}
\end{equation}
These features are aggregated using a single-layer GRU:
\begin{equation}
h^{(v)} = \text{GRU}_v\left(\{f^{(v)}_i\}_{i=1}^{16}\right), \quad h^{(v)} \in \mathbb{R}^{d_h}
\end{equation}

\subsubsection{Static Visual Representation from Image Frames}

In parallel, the same 64-frame window is used to uniformly select $16$ static frames $\{x_i^{(i)}\}_{i=1}^{16}$. Each frame is independently processed by the frozen \textbf{CLIP image encoder}:
\begin{equation}
f^{(i)}_j = \text{CLIP}_{\text{img}}(x_j^{(i)}), \quad f^{(i)}_j \in \mathbb{R}^{d_i}
\end{equation}
These image features are then passed through a GRU to produce a temporally-informed static representation:
\begin{equation}
h^{(i)} = \text{GRU}_i\left(\{f^{(i)}_j\}_{j=1}^{16}\right), \quad h^{(i)} \in \mathbb{R}^{d_h}
\end{equation}

\subsubsection{Semantic Representation from Generated Captions}

To encode high-level semantics, we uniformly select $4$ frames $\{x_k^{(t)}\}_{k=1}^{4}$ from the same 64-frame window. Each selected frame is captioned using the frozen \textbf{MiniGPT} model \cite{zhu2023minigpt4}, resulting in sentences $\{s_k\}_{k=1}^{4}$. These sentences are encoded via the frozen \textbf{CLIP text encoder}:
\begin{equation}
f^{(t)}_k = \text{CLIP}_{\text{text}}(s_k), \quad f^{(t)}_k \in \mathbb{R}^{d_t}
\end{equation}
A GRU is used to obtain a temporally-aware textual representation:
\begin{equation}
h^{(t)} = \text{GRU}_t\left(\{f^{(t)}_k\}_{k=1}^{4}\right), \quad h^{(t)} \in \mathbb{R}^{d_h}
\end{equation}

\subsection{Cross-Attention Fusion}

Given the modality-specific embeddings $h^{(v)}$, $h^{(i)}$, and $h^{(t)}$, we perform a two-stage bidirectional cross-attention process to integrate visual, spatial, and semantic information into a unified multimodal representation.

In the first stage, the image representation serves as the query to attend to the video modality:
\begin{equation}
z^{(i \leftarrow v)} = \text{Attention}\left(Q = h^{(i)},\; K = h^{(v)},\; V = h^{(v)}\right)
\end{equation}

In the second stage, the same image representation is used as the query to attend to the text modality:
\begin{equation}
z^{(i \leftarrow t)} = \text{Attention}\left(Q = h^{(i)},\; K = h^{(t)},\; V = h^{(t)}\right)
\end{equation}

We then concatenate the outputs from both stages to form the final fused multimodal embedding:
\begin{equation}
h^{(\text{fused})} = [z^{(i \leftarrow v)};\; z^{(i \leftarrow t)}] \in \mathbb{R}^{2d_h}
\end{equation}

This multimodal embedding serves as the input to the downstream task-specific head, such as a classifier for binary violence detection or a regressor for valence-arousal estimation.

\section{Experiment Result}

\subsection{Experimental Setup}

We evaluate our proposed multimodal architecture on two challenging benchmarks: the DVD dataset for violence classification and the Aff-Wild2 dataset for valence-arousal regression. Each dataset is split into 5 folds using the official validation protocols.

\textbf{Loss Function.} For the DVD dataset, we employ the \textit{Focal Loss}~\cite{lin2017focal} to address class imbalance in binary classification. For the Aff-Wild2 valence-arousal regression task, we use the \textit{Mean Squared Error (MSE)} as the primary loss.

\textbf{Optimization.} All models are trained using AdamW optimizer \cite{loshchilov2019decoupled} with a learning rate of $3 \times 10^{-4}$. Early stopping is employed based on validation performance. Batch size is set to 8 for both datasets.
\subsection{Results}

Table~\ref{tab:validation_results} reports the average validation results across five folds on each dataset.

\begin{table*}[htbp]
    \centering
    \caption{Validation Results of the Multimodal Model on DVD and Aff-Wild2 Datasets}
    \label{tab:validation_results}
    \begin{tabular}{lcccccc}
        \toprule
        \textbf{Dataset} & \textbf{Category} & \textbf{Fold 1} & \textbf{Fold 2} & \textbf{Fold 3} & \textbf{Fold 4} & \textbf{Fold 5} \\
        \midrule
        DVD & Multimodal & 0.9450 & 0.9442 & 0.9404 & 0.9380 & 0.9400 \\
        Aff-Wild2 & Multimodal & 0.8869 & 0.8915 & 0.8948 & 0.8880 & 0.8890 \\
        \bottomrule
    \end{tabular}
\end{table*}

Our multimodal model consistently achieves high performance on both tasks, with DVD reaching an average F1-score above 0.94 and Aff-Wild2 maintaining a CCC (concordance correlation coefficient) around 0.89. This demonstrates the robustness and generalizability of the proposed cross-attention-based fusion architecture.

\section{Conclusion}

In this paper, we proposed a novel multimodal framework for fine-grained video understanding that integrates motion, spatial, and semantic cues using a GRU-based architecture with cross-modal attention. The framework processes video, image, and text streams extracted from temporally aligned windows and effectively fuses them into a unified representation for downstream classification and regression tasks.

Experiments on two real-world benchmarks, DVD and Aff-Wild2, validate the effectiveness of our approach. Our model achieves state-of-the-art performance with only lightweight encoders and GRUs, demonstrating its potential for efficient deployment on real-time and edge applications. Future work may explore extending the model to other modalities (e.g., audio), incorporating transformer-based encoders, or enabling online adaptation during inference.

%Bibliography
\bibliographystyle{unsrt}  
\bibliography{references}

\end{document}